\documentclass[11pt]{article}
\usepackage{booktabs}
\usepackage{multirow}
\usepackage{array}
\usepackage[preprint]{acl}
\usepackage{pgfplots}
\usetikzlibrary{pgfplots.statistics}
\pgfplotsset{compat=1.18}
\usepackage{subcaption}
\usepackage{times}
\usepackage{latexsym}
\usepackage{listings}
\usepackage{xcolor}
\usepackage{mathtools}
\usepackage{subcaption}   % add to preamble
\usepackage{pgfplots}
\pgfplotsset{compat=1.18}

\usepackage[T1]{fontenc}
\usepackage[utf8]{inputenc}

\usepackage{microtype}

\usepackage{inconsolata}

\usepackage{graphicx}

\usepackage{amsmath}
\usepackage[dvipsnames]{xcolor}

\title{Geometry-Aware Hallucination Detection in Large Language Models}

\author{
  Bodla Krishna Vamshi \\
  University of Maryland, College Park \\
  \texttt{kbodla@umd.edu}
  \And
  Rohan Bhatnagar \\
  University of Maryland, College Park \\
  \texttt{rbhatna1@terpmail.umd.edu}
  \AND
  Haizhao Yang \\
  University of Maryland, College Park \\
  \texttt{hzyang@umd.edu}
}

\begin{document}
\maketitle
\begin{abstract}
Large language models (LLMs) frequently generate factually incorrect or unsupported content, commonly referred to as hallucinations. Prior work has explored decoding strategies, retrieval augmentation, and supervised fine-tuning for hallucination detection, while recent studies show that in-context learning (ICL) can substantially influence factual reliability. However, existing ICL demonstration selection methods often rely on surface-level similarity heuristics and exhibit limited robustness across tasks and models.

We propose GA-ICL, a geometry-aware demonstration sampling framework for selecting in-context demonstrations that leverages latent representations extracted from frozen LLMs. By jointly modeling local manifold structure and class-aware prototype geometry, GA-ICL selects demonstrations based on their proximity to learned prototypes rather than lexical or embedding similarity alone.

Across factual verification (FEVER) and hallucination detection (HaluEval) benchmarks, GA-ICL outperforms standard ICL selection baselines in the majority of evaluated settings, with particularly strong gains on dialogue and summarization tasks. The method remains robust under temperature perturbations and model variation, indicating improved stability compared to heuristic retrieval strategies. While lexical retrieval can remain competitive in certain question-answering regimes at smaller model scales, our results demonstrate that geometry-aware prototype selection provides a reliable and training-light approach for hallucination detection without modifying LLM parameters. Extended evaluations on Phi-14B and Qwen3-32B confirm that GA-ICL scales effectively to larger models, outperforming all compared baselines including on QA tasks where smaller models show boundary-condition limitations, offering a principled direction for improved ICL demonstration selection.
\end{abstract}

\section{Introduction}

\begin{figure}[t]
    \centering
    \includegraphics[width=\columnwidth]{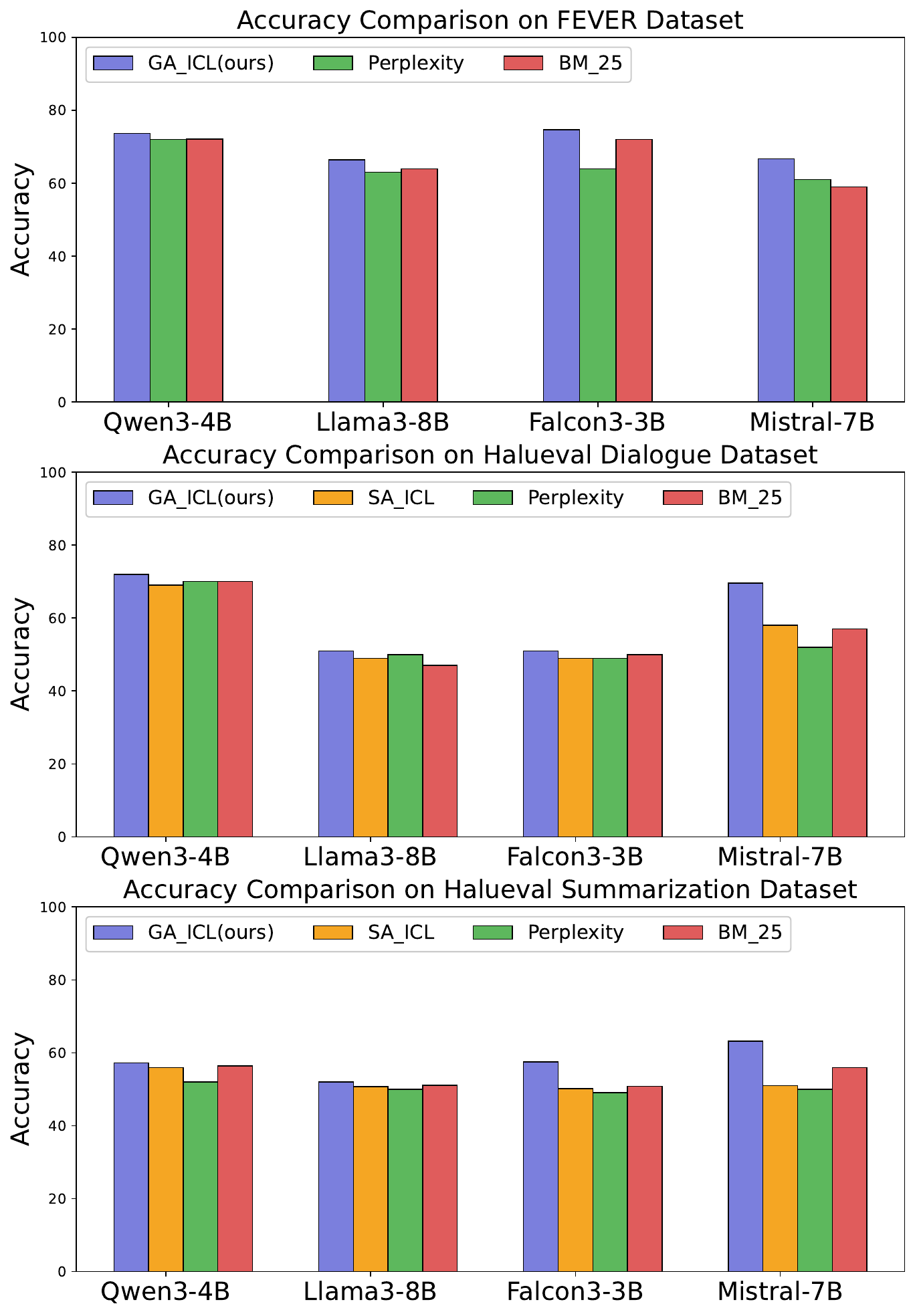}
    \caption{Accuracy comparisons of GA-ICL vs other methods}
    \label{fig:overview}
\end{figure}

The rapid scaling of pre-trained language models (PLMs) has led to the emergence of new capabilities~\cite{wei2022emergentabilitieslargelanguage}, particularly through in-context learning (ICL), where models perform downstream tasks by conditioning on prompts that include a small number of representative examples. Owing to its effectiveness and flexibility, ICL has become a widely adopted and efficient paradigm for utilizing PLMs. This paradigm has been successfully applied to a variety of tasks, including reasoning and code generation ~\cite{li2023largelanguagemodelawareincontext},~\cite{bodla2025protocodeprototypedriveninterpretabilitycode}, mathematical reasoning ~\cite{wu2025exampleshighlevelautomatedreasoning}, and question answering ~\cite{liu2023mmhqaiclmultimodalincontextlearning}, among others, without requiring explicit supervised fine-tuning.

In this study, we investigate the use of in-context learning (ICL) as a mechanism for hallucination detection in large language models. Unlike many existing methods that target hallucination detection through architectural modifications or specialized training and inference ~\cite{su-etal-2024-unsupervised} procedures, we examine whether leveraging ICL alone can provide comparable effectiveness for this task. Our experimental findings show that geometry-aware ICL yields gains over established baselines across a wide range of benchmarks and model configurations while attaining performance levels that are comparable to those achieved by supervised fine-tuning (Appendix ~\ref{sec:fine_tuning}). Importantly, these results are obtained with lower computational costs, highlighting the suitability of our approach for deployment in resource-constrained environments.

Existing ICL demonstration selection strategies largely rely on distance-based retrieval in a fixed embedding space, using lexical similarity, or heuristic ranking criteria. While effective in some settings, such approaches implicitly assume that the raw embedding geometry is well aligned with factual consistency, which may not hold across tasks and models. We argue that semantic similarity alone is insufficient for hallucination detection because embedding proximity does not necessarily correspond to factual-consistency similarity. GA-ICL addresses this mismatch by learning a retrieval geometry explicitly optimized for hallucination-discriminative demonstration selection. Unlike prior ICL selection methods that operate purely via distance-based retrieval in a fixed embedding space, our approach learns the geometry used for selection itself by explicitly modeling local manifolds and prototype structure. This shifts ICL selection from heuristic similarity ranking to geometry-based prototype sampling, which, to our knowledge, has been underexplored for hallucination-aware ICL retrieval.

\subsection{Contributions}

Our contributions are summarized as follows: 
% \begin{itemize}
%     \item \textbf{Manifold based prototype learning.} We introduce a framework to sample In-Context-Learning (ICL) demonstrations by combining \textit{piecewise-linear manifold learning} and \textit{proxy-anchor based metric learning} to construct a low-dimensional sampling space. This joint formulation ensures that the sampled prototypes are semantically discriminative. 
%     \item \textbf{Manifold grounded Hallucination Detection.} We demonstrate that manifold-based sampling enhances the ability of LLMs to identify factual inconsistencies without explicit parameter updates. Our method achieves performance competitive with supervised fine-tuning (SFT) (Appendix ~\ref{sec:fine_tuning}) while maintaining a substantially lower computational footprint. By leveraging pretrained models latent geometries, we present a scalable mechanism for hallucination detection that is robust across diverse architectures and settings. 
% \end{itemize}

\begin{itemize}
    \item We reformulate ICL demonstration selection for hallucination detection as a geometry-alignment problem rather than a purely heuristic similarity-retrieval problem. Unlike existing retrieval approaches that rely on fixed embedding similarity or lexical overlap, our method learns a task-specific latent geometry tailored for factual-consistency-aware retrieval. GA-ICL builds on the locally linear embedding tradition (~\cite{doi:10.1126/science.290.5500.2323}; ~\cite{NIPS2001_f106b7f9}), where high-dimensional data is approximated through local linear regions, and the prototype learning component follows the metric learning lineage of prototypical networks ~\cite{snell2017prototypicalnetworksfewshotlearning} and proxy anchor loss ~\cite{kim2020proxyanchorlossdeep}. The core contribution is the application of this combined paradigm to ICL demonstration selection a problem where no geometry-aware approach previously existed ~\cite{dong2024surveyincontextlearning}.
    
    \item We report an empirical finding that hallucination-discriminative information in frozen LLM representations occupies a substantially lower-dimensional subspace than the full embedding space a result not predicted by either manifold or metric learning theory in isolation. Reducing prototype dimensionality consistently improves detection accuracy across Qwen3-4B and Mistral-7B, and this compression effect is task-conditioned: gains are strongest on dialogue and summarization tasks where factual consistency requires semantic reasoning over long context, and weakest on QA tasks where explicit evidence snippets make lexical overlap near sufficient. This finding constitutes an empirical contribution beyond the method itself, revealing intrinsic geometric structure in how frozen LLMs encode hallucination-relevant information.
    
    \item We show that globally learned prototype retrieval achieves strong hallucination detection performance while substantially reducing inference-time retrieval overhead and memory consumption compared to dynamically retrieved ICL baselines.
    % \item We provide empirical evidence that hallucination discriminative information in frozen LLM representations occupies a substantially lower-dimensional subspace than the full embedding space. Reducing prototype dimensionality consistently improves detection performance across both Qwen3-4B and Mistral-7B, a result not predicted by either manifold or metric learning theory in isolation. This compression effect is task-conditioned: gains are strongest on dialogue and summarization where factual consistency requires semantic reasoning over long context, and weakest on QA tasks where explicit evidence snippets make lexical overlap near-sufficient. This finding constitutes an empirical contribution beyond the method itself, revealing intrinsic geometric structure in how LLMs encode hallucination-relevant information.
\end{itemize}

\section{Related Work}

\subsection{In-Context learning}
In-context learning (ICL) ~\cite{brown2020languagemodelsfewshotlearners} describes a paradigm in which language models solve tasks by relying on a limited set of example demonstrations provided at inference time, without modifying their internal parameters. Since these demonstrations are expressed directly in natural language, ICL enables an intuitive and interpretable mode of interaction with large language models (LLMs). This paradigm also aligns with key aspects of human cognition, particularly analogical reasoning, where prior examples inform decision-making in new situations ~\cite{10.1145/359038.359042}. Unlike conventional supervised learning approaches, ICL does not require additional training or fine-tuning, allowing models to flexibly generalize to novel tasks without incurring extra computational cost.

Building on these observations, numerous unsupervised methods have been proposed to identify informative demonstrations for in-context learning, as reviewed by~\cite{dong2024surveyincontextlearning}. One common line of work focuses on retrieving demonstrations that are closely related to the test input under predefined notions of similarity (~\cite{liu-etal-2022-makes}, ~\cite{tanwar-etal-2023-multilingual}, \citet{qin2024incontextlearningiterativedemonstration}). These approaches typically measure proximity using embedding-based distances, such as cosine similarity or Euclidean (L2) distance. Beyond similarity driven retrieval, prior studies have also explored alternative selection signals, including mutual information~\cite{sorensen-etal-2022-information} and perplexity~\cite{gonen-etal-2023-demystifying}, which enable effective prompt construction without requiring labeled supervision or access to a model’s internal parameters.

Although widely used in practice, generic retrieval based strategies often depend on heuristic design choices and can produce suboptimal in context examples due to the absence of explicit task level supervision. Importantly, these approaches typically assume that semantic similarity in the original embedding space is aligned with factual consistency, an assumption that may not hold for hallucination detection. To overcome these limitations, prior work has explored supervised retriever based methods for demonstration selection (~\cite{rubin-etal-2022-learning}, ~\cite{pmlr-v202-ye23c}, ~\cite{wang2024largelanguagemodelslatent}, ~\cite{zhang-etal-2022-active}). One representative approach is EPR proposed by~\cite{rubin-etal-2022-learning}, which employs a two stage training procedure to learn dense retrievers tailored for demonstration retrieval. Building on this idea,~\cite{li-etal-2023-unified} introduced a unified retrieval model capable of handling demonstration selection across diverse tasks. More recently,~\cite{mavromatis2023examplesannotateincontextlearning} proposed AdaICL, a model aware framework that leverages LLM generated pseudo labels and uncertainty estimates over unlabeled data to adaptively identify informative in context demonstrations.

\section{Method overview}

Given a dataset $D$, all textual fields are concatenated with a prompt template $T_p$ (Appendix ~\ref{sec:prompt_template}), resulting in a transformed dataset containing a single consolidated text field $F_t$ paired with its corresponding label $l$. The consolidated text field is passed through the corresponding LLM to obtain latent representations $Z$. The encoded latent representations $Z$ are obtained by taking the final transformer layer’s token hidden states and computing an attention mask weighted mean across all non padding tokens to produce a single fixed dimensional vector per input text. The extracted latent vector $Z$ is used for all the comparison baselines, including ours, to sample the ICL demonstrations with respect to each LLM.

As an initial step of our approach, we initialize a lightweight neural network $h_\theta$ (Table~\ref{tab:transformation_architecture}) that maps latent representations $Z$ to transformed representations $Z'$. In addition, we initialize two proxy vectors $\theta_q$, which serve as class representative prototypes within the proxy anchor loss, along with a momentum proxy vector $\theta_m$ that is updated according to the procedure outlined in Table~\ref{tab:algorithm_single_col}. Upon completion of training, the learned momentum proxy $\theta_m$ is used to retrieve the nearest transformed representations $Z'$, which are then selected as in context learning (ICL) demonstrations.

During training, for each mini batch $\mathcal{B} \subset D_{F_t}$, the latent representations $Z$ are passed through $h_\theta$ to obtain transformed representations $Z'$. As described in Section~\ref{sec:manifold_construction}, a manifold is then constructed for each batch based on $Z'$. The resulting manifold structure is used to compute the manifold point to point loss (Eq.~\ref{eq:manifold_loss}), while the transformed representations $Z'$ together with the proxy vectors $\theta_q$ are used to evaluate the proxy anchor loss (Eq.~\ref{eq:pca})

After computing both the manifold point to point loss and the proxy anchor loss, the overall training objective is formed as the sum of the two loss terms. The parameters of $h_\theta$ and the proxy vectors $\theta_q$ are then updated via gradient descent with respect to this combined loss, while the momentum proxy vector $\theta_m$ is updated following the procedure described in Table~\ref{tab:algorithm_single_col}. Upon completion of training, the learned momentum proxy $\theta_m$ is used to retrieve the nearest transformed representations $Z'$, which are selected as in context learning (ICL) demonstrations. The details of each parameter used and the training durations are clearly described in Appendix ~\ref{sec:imple_details}.

We emphasize that our approach is not training free. Although the underlying language model remains fully frozen, our method requires training a lightweight retrieval module in the form of a projection network $h\theta$ and associated prototype representations. This training phase is performed once per dataset and model configuration and is substantially less expensive than fine tuning the language model itself. We therefore position GA-ICL as a training light rather than training free approach, operating in contrast to standard ICL methods that rely solely on heuristic or unsupervised retrieval without learned parameters.

\begin{table}[t]
\centering
\footnotesize
\setlength{\tabcolsep}{3pt}
\renewcommand{\arraystretch}{1.15}
\begin{tabular}{p{\columnwidth}}
\hline
\textbf{Algorithm: Training and Sampling Overview} \\
\hline
\textbf{Input:} Dataset $D_{F_T}$ with labels $l$, momentum coefficient $\mu$ \\
\textbf{Output:} Prototype representation $\theta_q$ and selected ICL demonstrations \\
\hline
Initialize projection network $h_\theta$ and prototype $\theta_q$ \\
\hline
\textbf{for each batch} $\mathcal{B} \subset D_{F_t}$ \textbf{do} \\
\quad Obtain latent representations $Z \leftarrow \mathrm{LLM}(T)$ \\
\quad $Z' \leftarrow h_\theta(Z)$, Construct manifold structure for $Z'$ (Section ~\ref{sec:manifold_construction}) \\
\quad Compute loss:
$\mathcal{L} = \mathcal{L}_{\text{manifold}} + \mathcal{L}_{\text{PA}}$, eq. ~\ref{eq:manifold_loss} ,~\ref{eq:pca}\\
\quad Update parameters  of $h_\theta$ and $\theta_q$ via gradient descent \\
\quad Update Momentum proxy ~\cite{he2020momentumcontrastunsupervisedvisual}:
$\theta_m \leftarrow \mu \cdot \theta_m + (1 - \mu) \cdot \theta_q$ \\
\textbf{end for} \\
\hline
Select ICL demonstrations by sampling representations $Z'$ closest to $\theta_m$ \\
\hline
\end{tabular}
\caption{Overview of the Proposed GA-ICL Algorithm.}
\label{tab:algorithm_single_col}
\end{table}

\begin{table}[h!]
\caption{ $h_\theta$ Model Architecture}
\label{tab:transformation_architecture}
\centering
\begin{tabular}{@{}ll@{}}
\toprule
\textbf{Layer} & \textbf{Layer Parameters} \\
\midrule
Linear & (latent size $Z$, Prototype size $Z'$ ) \\
InstanceNorm1d & Prototype size $Z'$ \\
ReLU & - \\
\bottomrule
\end{tabular}
\end{table}

\subsection{Manifold Construction}
\label{sec:manifold_construction}
% Based on the Manifold hypothesis ~\cite{article}, we can assume that the encoded latent representations can be locally approximated into smaller chunks of linear regions. Our approach leverages this assumption ~\cite{bhatnagar2024piecewiselinearmanifoldsdeepmetric}, ~\cite{HOLIDAY2019419}, ~\cite{koronaki2023nonlineardimensionalityreductionnow}, ~\cite{sonday2010manifoldlearningtechniquesmodel} to identify representative prototypes that capture the characteristics of each action class. 

Based on the Manifold hypothesis ~\cite{article}, we can assume that the encoded latent representations can be locally approximated into smaller chunks of linear regions. This locally linear approximation follows the tradition of nonlinear dimensionality reduction methods such as Locally Linear Embedding ~\cite{doi:10.1126/science.290.5500.2323} and Laplacian Eigenmaps ~\cite{NIPS2001_f106b7f9}, which demonstrate that high-dimensional data concentrated near a lower-dimensional manifold can be faithfully represented through local linear patches. Our approach leverages this assumption ~\cite{bhatnagar2024piecewiselinearmanifoldsdeepmetric}, ~\cite{HOLIDAY2019419}, ~\cite{koronaki2023nonlineardimensionalityreductionnow}, ~\cite{sonday2010manifoldlearningtechniquesmodel} alongside prototype-based metric learning ~\cite{snell2017prototypicalnetworksfewshotlearning} to identify representative prototypes that capture the characteristics of each class.

\paragraph{Anchor Initialization and Local Clustering} To approximate the piecewise linear structure of the data, we aim to construct $m$-dimensional linear submanifolds around selected anchor points. Given a batch $B$ containing $N$ data points, we randomly select $n$ anchor points. For each anchor point $h_{\theta}(Z_i)$, we identify its nearest $m-1$ neighbors using Euclidean distance and form an initial neighborhood set $X_i$.

\paragraph{Manifold Expansion and Validation}
After initializing the local clusters, the manifold construction process proceeds iteratively by attempting to add the $m$-th nearest neighbor to $X_i$. After each addition, we compute the best fit $m$-dimensional submanifold using PCA to assess whether the members of $X_i$ can be reconstructed with quality above threshold $T\%$. If so, the $m$-th point is retained in $X_i$; otherwise it is excluded. This evaluation is repeated for subsequent neighbors $N(h_{\theta}(x_i))_j$ for $j \in \{ m + 1, \dots,k \}$, gradually constructing a local linear approximation of the manifold.

\paragraph{Submanifold Basis Extraction} The final set $X_i$ is comprised of all points in the anchor point's neighborhood that lie well within an $m$-dimensional linear submanifold. A basis for this submanifold is computed by applying PCA on $X_i$ and extracting the top $m$ eigenvectors. PCA is well suited for this task because it can effectively construct the lower-dimensional manifolds for locally linear regions. 

\subsection{Training Objectives}

\label{sec:training_loss}
% Our approach combines two loss functions: a modified version of proxy anchor loss for metric learning and a manifold-based point-to-point similarity loss.
\paragraph{Proxy Anchor Loss:}
We use a modified version of proxy anchor loss ~\cite{kim2020proxyanchorlossdeep} with Euclidean distance instead of cosine similarity:
%----equation proxy anchor loss ------------------------

\begin{equation}
\label{eq:pca}
\scalebox{0.75}{$
\begin{aligned}
\mathcal{L}_{\text{PA}} &=
\frac{1}{|\Theta_+|}
\sum_{\theta_q \in \Theta_+}
\log\!\Bigg(
1 + \sum_{\mathclap{Z' \in \mathcal{Z'}_{\theta_q}^+}}
\exp\!\Big(
-\alpha \big(\|h_\theta(Z)-\theta_q\|_2-\epsilon\big)
\Big)
\Bigg) \\
&\quad+
\frac{1}{|\Theta|}
\sum_{\theta_q \in \Theta}
\log\!\Bigg(
1 + \sum_{\mathclap{Z' \in \mathcal{Z'}_{\theta_q}^-}}
\exp\!\Big(
\alpha \big(\|h_\theta(Z)-\theta_q\|_2-\epsilon\big)
\Big)
\Bigg)
\end{aligned}
$}
\end{equation}

Let $\Theta$ represent the collection of all proxy vectors, where each proxy $\theta_q \in \Theta$ functions as a class-level representative in the embedding space. From this set, $\Theta_+ \subseteq \Theta$ denotes the subset of proxies that are associated with at least one positive sample within the current mini-batch $B$. For any selected proxy $\theta_q$, the latent embeddings $\mathcal{Z'}$ contained in $B$ (with $Z' \in \mathcal{Z'}$) are divided into two disjoint groups: the positive set $\mathcal{Z'}{\theta_q}^+$, consisting of embeddings from the same class as $\theta_q$, and the negative set $\mathcal{Z'}{\theta_q}^- = \mathcal{Z'} \setminus \mathcal{Z'}_{\theta_q}^+$, which contains all remaining embeddings. The temperature parameter $\alpha$ regulates the concentration of the optimization process: larger values emphasize hard positive and negative pairs by sharpening gradients, whereas smaller values yield smoother updates by distributing influence more uniformly across pairs. The margin parameter $\epsilon$ introduces an explicit separation constraint, ensuring that embeddings from the same class are pulled closer to their corresponding proxies while embeddings from different classes are pushed beyond a minimum distance threshold.

\paragraph{Manifold Point-to-Point Loss:}
\label{para:manifold_sim}
This loss helps in estimating the point to point similarities preserving the geometric structure:
% \begin{equation}
% \label{eq:manifold_loss}
% \small
% \mathcal{L}_{\text{manifold}} = \sum_{i,j} \left(\delta \cdot (1 - s(z_i, z_j)) - \|h_\theta(z_i) - h_\theta(z_j)\|_2\right)^2
% \end{equation}
% where \( s(z_i, z_j) \) is the manifold-based similarity computed as
% $s(z_i, z_j) = \frac{s'(z_i, z_j) + s'(z_j, z_i)}{2}$ with \( s'(z_i, z_j) = \alpha(z_i, z_j) \cdot \beta(z_i, z_j) \), where
% $\alpha(z_i, z_j) = \frac{1}{\left(1 + o(z_i, z_j)^2\right)^{N_\alpha}}$ and $\beta(z_i, z_j)  = \frac{1}{\left(1 + p(z_i, z_j)\right)^{N_\beta}}$. 

\begin{equation}
\label{eq:manifold_loss}
\small
\mathcal{L}_{\text{manifold}} = \sum_{i,j} \left(\delta \cdot (1 - s(Z'_i, Z'_j)) - \|h_\theta(Z_i) - h_\theta(Z_j)\|_2\right)^2
\end{equation}

\noindent where \( s(Z'_i, Z'_j) \) is the manifold similarity computed as:
\begin{equation*}
\small
s(Z'_i, Z'_j) = \frac{s'(Z'_i, Z'_j) + s'(Z'_j, Z'_i)}{2}
\end{equation*}

\noindent with \( s'(Z'_i, Z'_j) = \alpha(Z'_i, Z'_j) \cdot \beta(Z'_i, Z'_j) \), where:
\begin{equation*}
\small
\begin{aligned}
\alpha(Z'_i, Z'_j) &= \frac{1}{\left(1 + o(Z'_i, Z'_j)^2\right)^{N_\alpha}} \\
\beta(Z'_i, Z'_j)  &= \frac{1}{\left(1 + p(Z'_i, Z'_j)\right)^{N_\beta}}
\end{aligned}
\end{equation*}

In \eqref{eq:manifold_loss}, \( h_\theta \) denotes a lightweight neural transformation whose architecture is detailed in Table~\ref{tab:transformation_architecture}, and \( \delta \) is a scaling constant that sets an upper bound on the separation between dissimilar samples. This objective aligns Euclidean distances in the transformed embedding space with manifold-induced dissimilarities \( 1 - s(Z'_i, Z'_j) \), thereby encouraging the learned metric to conform to the intrinsic geometry of the data manifold. The term \( o(Z'_i, Z'_j) \) measures the orthogonal distance from point \( z_i \) to the manifold associated with \( Z'_j \), while \( p(Z'_i, Z'_j) \) captures the distance between \( Z'_j \) and the projection of \( z_i \) onto that manifold. The hyperparameters \( N_\alpha \) and \( N_\beta \) regulate the rate at which similarity attenuates with increasing distance, with the constraint \( N_\alpha > N_\beta \) ensuring that off-manifold points experience a steeper decay in similarity than points residing on the same manifold.

\paragraph{Distance Calculation.}
For each embedding pair $(Z'_i, Z'_j)$, the distances $o(Z'_i, Z'_j)$ and $p(Z'_i, Z'_j)$ are computed using the local manifold basis $P_j$ associated with point $Z'_j$. The projection of $Z'_i$ onto the manifold spanned by $P_j$ is defined as $\text{proj}_{P_j}(Z'_i) = Z'_j + \sum_{k} \langle Z'_i - Z'_j, v_k \rangle v_k$, where $\{v_k\}$ denotes the basis vectors of $P_j$. The orthogonal distance is then given by $o(Z'_i, Z'_j) = \| Z'_i - \text{proj}_{P_j}(Z'_i) \|_2$, while the projected distance along the manifold is defined as $p(Z'_i, Z'_j) = \| \text{proj}_{P_j}(Z'_i) - Z'_j \|_2$. This procedure is applied to all point pairs, enabling the loss to capture the local geometric structure of the data manifold.

\section{Experiments}
\label{sec:experiments}
We focus on hallucination detection, where the model predicts factual consistency labels, rather than modifying generation to reduce hallucinations. We conduct experiments across six large language models on hallucination detection (HaluEval) and factual verification (FEVER) tasks to evaluate our sampling approach. We follow prior work in in-context learning and report single run accuracy since demonstration selection is deterministic given fixed representations and prompts. Demonstration selection, prompt construction, and decoding are all fixed across runs, and no stochastic sampling or retriever randomness is introduced. Our method (\textbf{GA-ICL})  attains higher accuracy relative to baseline methods in most of the scenarios. Additionally, lower perplexity scores compared to BM25 and self adaptive ICL sampling suggest reduced predictive uncertainty. To assess robustness to decoding stochasticity, we examine accuracy across different sampling temperatures. For the Qwen3-4B and Falcon3-3B models, our method shows limited sensitivity to temperature changes, in contrast to BM25 and self adaptive ICL based sampling methods, indicating more concentrated predictive distributions under typical inference settings. Our ablation study across all model parameters indicates that the prototype size $Z'$
has the largest impact on performance. In particular, when the original latent dimension $Z$ is substantially reduced to $Z'$, we observe consistent changes in performance, suggesting the presence of manifold learning effects induced by the reduced representation capacity.

\subsection{Dataset}

We conduct experiments across four distinct tasks using six different large language models. The FEVER ~\cite{thorne-etal-2018-fever} dataset is a widely used benchmark for evaluating a model’s ability to assess the factual correctness of a given sentence. It consists of two splits (train and test), where each instance contains a claim paired with a label indicating whether the statement is supported or refuted. The HaluEval ~\cite{li-etal-2023-halueval} dataset is comprised of 5,000 general user queries paired with ChatGPT-generated responses, along with 30,000 task specific examples spanning three tasks: question answering, knowledge grounded dialogue conversation, and summarization. In the HaluEval question answering setting, the model is provided with a question, a supporting knowledge snippet, and a candidate answer, and is required to determine whether the answer is hallucinated. The dialogue and summarization subsets follow a similar format.

\subsection{Evaluation details \& Results}
\label{sec:details_and_results}

All sampling strategies, including our proposed method, were evaluated under a two shot in context learning (ICL) setting, where each method selected exactly two demonstrations per input across all models and tasks. This setup ensures that performance differences arise from the quality of the selected demonstrations rather than the quantity of contextual information provided to the model.

For the FEVER dataset, which includes predefined training and test splits, the training split was used exclusively for sampling ICL demonstrations, while evaluation was conducted on the held out test split. For the HaluEval tasks summarization, question answering, and dialogue the dataset does not provide explicit splits. In these cases, demonstrations were sampled directly from the dataset and subsequently removed prior to evaluation to prevent data leakage. All methods followed the same procedure to maintain consistency across comparisons.

Table ~\ref{tab:main_results} summarizes the performance of all methods across six language models and four tasks. Across the majority of settings, our manifold based ICL sampling approach (GA-ICL) achieves higher accuracy than baseline retrieval and selection strategies, including BM25, perplexity based selection, clustering, and self-adaptive ICL (SA-ICL). GA-ICL improves hallucination detection accuracy across most models and tasks, with particularly strong gains in dialogue and summarization settings. To evaluate the scalability of our method on larger language models, we compared it against several baselines using Phi-14B and Qwen3-32B. As shown in Table ~\ref{tab:large_model_results}, the results demonstrate that our method generalizes effectively to larger models.

% These improvements are observed consistently across both factual verification (FEVER) and hallucination detection (HaluEval) tasks, suggesting that the proposed sampling strategy generalizes across task formats and model architectures.

GA-ICL improves performance across models and tasks in the majority of settings. On Mistral-7B, the method achieves large gains on dialogue and summarization hallucination detection, with absolute improvements of up to 9.9 accuracy points over standard retrieval baselines (Table ~\ref{tab:main_results}). For Falcon3-3B, GA-ICL similarly yields clear improvements on the summarization task, where baseline selection strategies show limited effectiveness. In contrast, recent adaptive ICL selection methods generally report more modest gains, typically on the order of 1–2 percentage points, in comparable settings, highlighting the impact of effective demonstration selection for hallucination detection tasks. We note that in certain question-answering settings, lexical retrieval methods such as BM25 outperform GA-ICL, suggesting that surface level overlap can remain a strong signal when factual correctness is tightly coupled to explicit evidence matching.

To further examine the effect of the number of demonstrations, we conducted an ablation study varying the ICL shot count for the Qwen3-4B and Falcon3-3B models. As shown in Figures ~\ref{fig:qwen3_accuracy} and ~\ref{fig:falcon3_accuracy}, performance improves as the number of demonstrations increases up to approximately ten shots, after which gains begin to saturate. This trend is consistent across both dialogue and question answering tasks, indicating diminishing returns from additional demonstrations beyond this point.

% During the hyperparameter tuning, we found that the impact of the prototype dimensionality $Z'$ on LLM performance was more influential than that of the others (Fig ~\ref{fig:hyper_param_overall}). Figures ~\ref{fig:mistral_dialogue_latent_size_accuracy} and ~\ref{fig:qwen_latent_size_accuracy} show accuracy as a function of prototype size for the Qwen3-4B and Mistral-7B models. Across both models, reducing the prototype dimensionality leads to consistent performance improvements. This trend suggests that projecting representations into a more compact space improves the quality of selected demonstrations, likely by suppressing task irrelevant variation while preserving the information necessary for hallucination detection. This observation suggests that hallucination-discriminative information may occupy a lower-dimensional latent structure than the original embedding space, indicating that compressed retrieval geometries can improve factual-consistency-aware demonstration selection.

\textbf{Prototype dimensionality as a window into LLM hallucination geometry}. Across both Qwen3-4B and Mistral-7B, reducing the prototype space to $Z'$ from the full LLM hidden dimension consistently improves hallucination detection accuracy (Figures ~\ref{fig:mistral_dialogue_latent_size_accuracy}–~\ref{fig:qwen_latent_size_accuracy}). These trends are unlikely to arise solely from regularization effects: it is task-conditioned gains are strongest on dialogue and summarization, where factual consistency requires reasoning over long context, and weakest on QA, where explicit evidence snippets make lexical overlap near-sufficient. We interpret this as evidence that hallucination-relevant information may reside in a lower-dimensional latent subspace within frozen LLMs, and that compact retrieval geometries can better isolate it.

Although our method includes an additional training stage to learn prototype samples, this overhead is compensated during inference. As shown in Appendix \ref{sec: inference_table}, GA-ICL achieves the lowest inference time and memory consumption among all compared baselines. This efficiency arises from its global sampling strategy, where prototypes are selected once for the entire dataset and reused uniformly across all test instances, avoiding the repeated per-query retrieval cost incurred by dynamically sampling baselines.

Finally, we note that these trends are consistent across different model families and tasks, indicating that the observed improvements are not tied to a specific architecture or dataset. Collectively, these results demonstrate that the proposed manifold based sampling strategy provides a reliable and effective mechanism for selecting ICL demonstrations in hallucination detection settings. All reported results are obtained with the decoding temperature fixed at 0, making both demonstration selection and generation fully deterministic. Following prior work on deterministic in context learning, we therefore report single run accuracy. Robustness is instead assessed via temperature perturbation analysis, as reported in Section ~\ref{sec:temp_analysis}.

\begin{table*}[h]
\centering
\small
\renewcommand{\arraystretch}{1.1}
\setlength{\tabcolsep}{2pt}
\begin{tabular}{ll cccc cc cc}
\toprule
\textbf{Model} & \textbf{Task} & \textbf{KNN} & \textbf{Clustering} & \textbf{BM25} & \textbf{Perplexity} & \textbf{SA-ICL} & \textbf{AdaICL} & \textbf{Top-k+Con-E} & \textbf{GA-ICL (Ours)} \\
\midrule
\multirow{4}{*}{Qwen3-4B} 
 & HaluEval Dialogue & 14.7 & 49.9 & 70.0 & 70.0 & 69.0 & 67.8 & 65.0 & \textbf{72.0}  \\
 & HaluEval Summ.    & 53.4 & 54.9 & 56.4 & 52.0 & 56.0 & 53.3 & 52.7 & \textbf{57.2}  \\
 & HaluEval QA       & 12.7 & 50.0 & 69.0 & \textbf{74.0} & 71.0 & 68.7 & 64.3 & 70.5  \\
 & FEVER             & 72.2 & 70.5 & 72.1 & 72.0 & --   & --   & --   & \textbf{73.7}  \\
\midrule
\multirow{4}{*}{Llama3-8B} 
 & HaluEval Dialogue & 11.1 & 50.0 & 47.0 & 50.0 & 49.0 & 56.4 & 49.0 & \textbf{51.0}  \\
 & HaluEval Summ.    & 51.7 & 50.1 & 51.1 & 50.0 & 50.7 & 51.9 & 49.4 & \textbf{52.0}  \\
 & HaluEval QA       & 7.0  & 50.0 & 63.0 & 62.0 & 65.0 & 61.9 & 48.1 & \textbf{69.0}  \\
 & FEVER             & 65.5 & 62.8 & 64.0 & 63.0 & --   & --   & --   & \textbf{66.4}  \\
\midrule
\multirow{4}{*}{Falcon-3-3B} 
 & HaluEval Dialogue & 5.9  & 50.0 & 50.0 & 49.0 & 49.0 & 50.4 & 50.0 & \textbf{51.0}  \\
 & HaluEval Summ.    & 50.4 & 51.8 & 50.8 & 49.1 & 50.2 & 54.7 & 51.9 & \textbf{57.5}  \\
 & HaluEval QA       & 5.8  & 50.0 & 49.0 & 50.0 & 50.0 & 50.2 & 49.3 & \textbf{51.0}  \\
 & FEVER             & 73.8 & 71.8 & 72.0 & 64.0 & --   & --   & --   & \textbf{74.7}  \\
\midrule
\multirow{4}{*}{GPT-Neo-2.7B} 
 & HaluEval Dialogue & --   & 49.9 & 50.0 & 48.0 & 50.0 & 50.1 & 49.0 & \textbf{52.0}  \\
 & HaluEval QA       & --   & 50.0 & 49.0 & 45.0 & 49.0 & 48.6 & 49.4 & \textbf{52.0}  \\
 & FEVER             & 43.0 & 50.0 & 49.0 & 49.0 & --   & --   & --   & \textbf{50.2}  \\
\midrule
\multirow{4}{*}{Mistral-7B} 
 & HaluEval Dialogue & 59.7 & 49.9 & 57.0 & 52.0 & 58.0 & 56.4 & 57.0 & \textbf{69.6}  \\
 & HaluEval Summ.    & 39.9 & 57.2 & 56.0 & 50.0 & 51.0 & 51.2 & 54.9 & \textbf{63.2}  \\
 & HaluEval QA       & 62.5 & 50.0 & \textbf{66.0} & 52.0 & 62.0 & 59.3 & 58.0 & 62.0  \\
 & FEVER             & 64.1 & 64.7 & 59.0 & 61.0 & --   & --   & --   & \textbf{66.7}  \\
\midrule
\multirow{4}{*}{Vicuna-7B} 
 & HaluEval Dialogue & 16.4 & 49.9 & 49.0 & 50.0 & 50.0 & 49.5 & 49.0 & \textbf{51.0}  \\
 & HaluEval Summ.    & 50.0 & 50.7 & 26.0 & 0.0  & 39.0 & 52.7 & 52.8 & \textbf{56.3}  \\
 & HaluEval QA       & 26.8 & 50.0 & 50.0 & 50.0 & 50.0 & 49.3 & 49.0 & \textbf{51.0}  \\
 & FEVER             & 65.5 & 61.3 & 67.0 & 66.3 & --   & --   & --   & \textbf{71.1}  \\
\bottomrule
\end{tabular}
\caption{Performance comparison (Accuracy) of GA-ICL against baselines. Missing values (–) indicate tasks where the baseline was not applicable or data was unavailable for that specific model configuration. Result uses optimized prototype dimensionality identified via ablation study (Figure ~\ref{fig:mistral_qa_latent_size_accuracy}); default hyperparameters yield 53.0.}
\label{tab:main_results}
\end{table*}

\begin{table*}[t]
\centering
\footnotesize
\renewcommand{\arraystretch}{1.0}
\setlength{\tabcolsep}{4pt}

\begin{tabular}{ll ccccc}
\toprule
\textbf{Model} & \textbf{Task} & \textbf{Perplexity} & \textbf{SA-ICL} & \textbf{AdaICL} & \textbf{Top-k+Con-E} & \textbf{GA-ICL (Ours)} \\
\midrule
\multirow{3}{*}{Phi-14B}
 & HaluEval Dialogue & 70.3 & 72.0 & 71.4 & 65.3 & \textbf{75.9} \\
 & HaluEval Summ.    & 60.3 & 59.0 & 62.1 & 61.9 & \textbf{65.4} \\
 & HaluEval QA       & 76.7 & 78.3 & 73.0 & 76.7 & \textbf{81.2} \\
\midrule
\multirow{3}{*}{Qwen3-32B}
 & HaluEval Dialogue & 80.1 & 82.0 & 81.0 & 79.3 & \textbf{83.5} \\
 & HaluEval Summ.    & 63.5 & 60.9 & 60.0 & 64.8 & \textbf{67.3} \\
 & HaluEval QA       & 66.5 & 67.0 & 65.4 & 67.5 & \textbf{71.7} \\
\bottomrule
\end{tabular}

\caption{Accuracy comparison of GA-ICL on large scale models.}
\label{tab:large_model_results}
\end{table*}

% \begin{table}[h!]
% \centering
% \footnotesize
% \renewcommand{\arraystretch}{1.8}
% \setlength{\tabcolsep}{3pt}
% \resizebox{\columnwidth}{!}{%
% \begin{tabular}{lcccccc}
% \noalign{\hrule height 1.2pt}
% \textbf{Method $\downarrow$} & \multicolumn{6}{c}{\textbf{Model $\rightarrow$}} \\
% \cline{2-7}
%  & \textbf{Qwen3-4B}
%  & \textbf{Llama3-8B}
%  & \textbf{Falcon-3-3B}
%  & \textbf{GPT-Neo-2.7B}
%  & \textbf{Vicuna-7B}
%  & \textbf{Mistral-7B} \\
% \hline
% KNN       & 12.7 & 7.0 & 5.8 & -    & 26.8 & 62.5 \\
% \hline
% Clustering        & 50.0 & 50.0 & 50.0 & 50.0 & 50.0 & 50.0 \\
% \hline
% BM25             & 69.0 & 63.0 & 49.0 & 49.0 & 50.0 & \textbf{66.0} \\
% \hline
% Perplexity       & \textbf{74.0} & 62.0 & 50.0 & 45.0 & 50.0 & 52.0 \\
% \hline
% SA-ICL           & 71.0 & 65.0 & 50.0 & 49.0 & 50.0 & 62.0 \\
% \hline
% GA-ICL(ours)       & 70.1 & \textbf{69.0} & \textbf{51.0} & \textbf{52.0} & \textbf{51.0} & 53.0 \\
% \noalign{\hrule height 1.2pt}
% \end{tabular}
% }
% \caption{Accuracy for the Q\&A task on the HaluEval dataset.}
% \label{tab:halueval_qa}
% \end{table}

\subsection{Effect of Temperature and Perplexity based evaluation}
\label{sec:temp_analysis}
As an additional sanity check, we examined the effect of generation temperature on model behavior by varying the temperature from 0 to 1 in increments of 0.1. We compared the resulting output variability of our method with that of \textbf{BM25} and \textbf{SA-ICL} on the HaluEval question answering and dialogue tasks Fig ~\ref{fig:diag_qwen}, ~\ref{fig:diag_falcon}, ~\ref{fig:qa_qwen}, ~\ref{fig:qa_falcon}. Across this range, our approach exhibited more stable behavior relative to the comparison methods. Specifically, as the temperature increased from $T=0$ to $T=1$, the variation in model outputs remained limited, suggesting that the sampling procedure is comparatively less sensitive to temperature changes. 

This pattern is consistent with a more concentrated token probability distribution during generation. In addition to accuracy, we report average perplexity scores under fixed decoding conditions (temperature = 0) as a complementary diagnostic signal. Perplexity here is not interpreted as a calibrated measure of model uncertainty, but rather as an indicator of how well the selected in-context demonstrations align with the model’s internal likelihood structure. Lower perplexity suggests that the resulting prompts induce outputs that are more consistent with the model’s learned distribution, which empirically correlates with improved hallucination detection performance in our setting. Importantly, we treat perplexity as an observational metric and do not claim a causal relationship between perplexity reduction and factual correctness.

% \begin{table}[h!]
% \centering
% \renewcommand{\arraystretch}{1.6}
% \setlength{\extrarowheight}{2pt}
% \setlength{\tabcolsep}{4pt}
% \resizebox{\columnwidth}{!}{%
% {\fontsize{9}{11}\selectfont
% \begin{tabular}{l|cccccc}
% \hline
% \textbf{Method} & \textbf{Qwen3-4B} & \textbf{Llama3-8B} & \textbf{Falcon-3-3B} & \textbf{GPT-Neo-2.7B} & \textbf{Vicuna-7B} & \textbf{Mistral-7B} \\
% \hline
% SA-ICL & $14.1 \pm 3.5$ & $6.8 \pm 1.2$ & $7.4 \pm 1.3$ & $9.2 \pm 1.6$ & $7.1 \pm 1.2$ & $5.1 \pm 0.7$ \\
% BM25 & $16.1 \pm 2.1$ & $7.5 \pm 0.5$ & $8.1 \pm 0.8$ & $13.2 \pm 1.3$ & $6.8 \pm 0.5$ & $5.8 \pm 0.4$ \\
% GA-ICL (ours) & \textbf{$12.8 \pm 3.0$} & \textbf{$6.3 \pm 1.0$} & \textbf{$7.0 \pm 1.2$} & \textbf{$8.9 \pm 1.5$} & \textbf{$6.5 \pm 1.0$} & \textbf{$4.8 \pm 0.6$} \\
% \hline
% \end{tabular}
% }}
% \caption{Perplexity of models averaged across samples from the HaluEval dialogue task.}
% \label{tab:comparison}
% \end{table}

% --- Preamble (make sure you have these) ---

% --- Figure ---
\begin{figure}[h!]
\centering
\begin{tikzpicture}
\begin{axis}[
    ybar,
    bar width=7pt,
    width=\columnwidth,
    height=7cm,
    ylabel={Perplexity},
    xlabel={Models},
    symbolic x coords={Qwen3-4B, {}, Llama3-8B, {}, Falcon-3-3B, {}, GPT-Neo-2.7B, {}, Vicuna-7B, {}, Mistral-7B},
    xtick={Qwen3-4B, Llama3-8B, Falcon-3-3B, GPT-Neo-2.7B, Vicuna-7B, Mistral-7B},
    x tick label style={rotate=45, anchor=east, font=\small},
    legend style={at={(0.97,0.97)}, anchor=north east, font=\small, fill=white, fill opacity=0.8, draw opacity=1, text opacity=1, draw=black},
    ymajorgrids=true,
    grid style=dashed,
    enlarge x limits=0.08,
    ymin=0, ymax=22,          % bump a bit to avoid clipping
    clip=false,               % prevents error bars from being clipped
]

% SA-ICL
\addplot+[
  fill=blue!60, draw=blue!80!black,
  error bars/.cd,
    y dir=both,
    y explicit,
    error bar style={line width=0.6pt},
    error mark options={rotate=90, mark size=2pt, line width=0.6pt}
]
coordinates {
    (Qwen3-4B, 14.1) +- (0, 3.5)
    ({}, 0)
    (Llama3-8B, 6.8) +- (0, 1.2)
    ({}, 0)
    (Falcon-3-3B, 7.4) +- (0, 1.3)
    ({}, 0)
    (GPT-Neo-2.7B, 9.2) +- (0, 1.6)
    ({}, 0)
    (Vicuna-7B, 7.1) +- (0, 1.2)
    ({}, 0)
    (Mistral-7B, 5.1) +- (0, 0.7)
};

% BM25
\addplot+[
  fill=red!60, draw=red!80!black,
  error bars/.cd,
    y dir=both,
    y explicit,
    error bar style={line width=0.6pt},
    error mark options={rotate=90, mark size=2pt, line width=0.6pt}
]
coordinates {
    (Qwen3-4B, 16.1) +- (0, 2.1)
    ({}, 0)
    (Llama3-8B, 7.5) +- (0, 0.5)
    ({}, 0)
    (Falcon-3-3B, 8.1) +- (0, 0.8)
    ({}, 0)
    (GPT-Neo-2.7B, 13.2) +- (0, 1.3)
    ({}, 0)
    (Vicuna-7B, 6.8) +- (0, 0.5)
    ({}, 0)
    (Mistral-7B, 5.8) +- (0, 0.4)
};

% GA-ICL (ours)
\addplot+[
  fill=green!60, draw=green!80!black,
  error bars/.cd,
    y dir=both,
    y explicit,
    error bar style={line width=0.6pt},
    error mark options={rotate=90, mark size=2pt, line width=0.6pt}
]
coordinates {
    (Qwen3-4B, 12.8) +- (0, 3.0)
    ({}, 0)
    (Llama3-8B, 6.3) +- (0, 1.0)
    ({}, 0)
    (Falcon-3-3B, 7.0) +- (0, 1.2)
    ({}, 0)
    (GPT-Neo-2.7B, 8.9) +- (0, 1.5)
    ({}, 0)
    (Vicuna-7B, 6.5) +- (0, 1.0)
    ({}, 0)
    (Mistral-7B, 4.8) +- (0, 0.6)
};

\legend{SA-ICL, BM25, GA-ICL (ours)}
\end{axis}
\end{tikzpicture}
\caption{Perplexity comparison across models on HaluEval dialogue task. Lower perplexity indicates better performance.}
\label{fig:perplexity_comparison}
\end{figure}
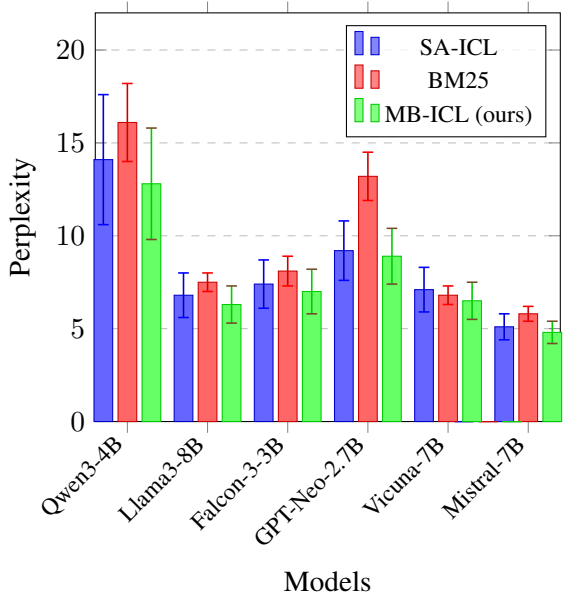

% --- Preamble (needed) ---

% --- Figure: Perplexity (QA task) with error bars from Table ---
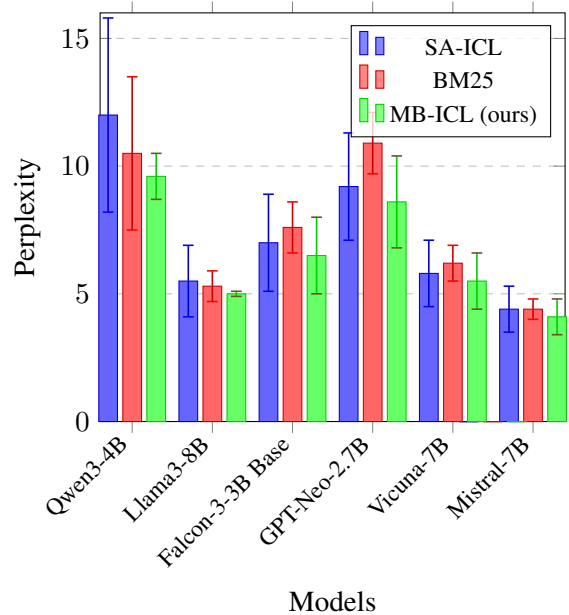
\begin{figure}[h!]
\centering
\begin{tikzpicture}
\begin{axis}[
    ybar,
    bar width=7pt,
    width=\columnwidth,
    height=7cm,
    ylabel={Perplexity},
    xlabel={Models},
    symbolic x coords={Qwen3-4B, {}, Llama3-8B, {}, Falcon-3-3B Base, {}, GPT-Neo-2.7B, {}, Vicuna-7B, {}, Mistral-7B},
    xtick={Qwen3-4B, Llama3-8B, Falcon-3-3B Base, GPT-Neo-2.7B, Vicuna-7B, Mistral-7B},
    x tick label style={rotate=45, anchor=east, font=\small},
    legend style={at={(0.97,0.97)}, anchor=north east, font=\small, fill=white, fill opacity=0.8, draw opacity=1, text opacity=1, draw=black},
    ymajorgrids=true,
    grid style=dashed,
    enlarge x limits=0.08,
    ymin=0, ymax=16,  % enough headroom for mean+std
    clip=false,
]

% SA-ICL
\addplot+[
  fill=blue!60, draw=blue!80!black,
  error bars/.cd,
    y dir=both,
    y explicit,
    error bar style={line width=0.6pt},
    error mark options={rotate=90, mark size=2pt, line width=0.6pt}
]
coordinates {
    (Qwen3-4B, 12.0) +- (0, 3.8)
    ({}, 0)
    (Llama3-8B, 5.5) +- (0, 1.4)
    ({}, 0)
    (Falcon-3-3B Base, 7.0) +- (0, 1.9)
    ({}, 0)
    (GPT-Neo-2.7B, 9.2) +- (0, 2.1)
    ({}, 0)
    (Vicuna-7B, 5.8) +- (0, 1.3)
    ({}, 0)
    (Mistral-7B, 4.4) +- (0, 0.9)
};

% BM25
\addplot+[
  fill=red!60, draw=red!80!black,
  error bars/.cd,
    y dir=both,
    y explicit,
    error bar style={line width=0.6pt},
    error mark options={rotate=90, mark size=2pt, line width=0.6pt}
]
coordinates {
    (Qwen3-4B, 10.5) +- (0, 3.0)
    ({}, 0)
    (Llama3-8B, 5.3) +- (0, 0.6)
    ({}, 0)
    (Falcon-3-3B Base, 7.6) +- (0, 1.0)
    ({}, 0)
    (GPT-Neo-2.7B, 10.9) +- (0, 1.2)
    ({}, 0)
    (Vicuna-7B, 6.2) +- (0, 0.7)
    ({}, 0)
    (Mistral-7B, 4.4) +- (0, 0.4)
};

% GA-ICL (ours)
\addplot+[
  fill=green!60, draw=green!80!black,
  error bars/.cd,
    y dir=both,
    y explicit,
    error bar style={line width=0.6pt},
    error mark options={rotate=90, mark size=2pt, line width=0.6pt}
]
coordinates {
    (Qwen3-4B, 9.6) +- (0, 0.9)
    ({}, 0)
    (Llama3-8B, 5.0) +- (0, 0.1)
    ({}, 0)
    (Falcon-3-3B Base, 6.5) +- (0, 1.5)
    ({}, 0)
    (GPT-Neo-2.7B, 8.6) +- (0, 1.8)
    ({}, 0)
    (Vicuna-7B, 5.5) +- (0, 1.1)
    ({}, 0)
    (Mistral-7B, 4.1) +- (0, 0.7)
};

\legend{SA-ICL, BM25, GA-ICL (ours)}
\end{axis}
\end{tikzpicture}
\caption{Perplexity comparison across models on HaluEval QA task. Lower perplexity indicates better performance.}
\label{fig:perplexity_comparison_qa}
\end{figure}

\subsection{Failure analysis and Limitations}

GA-ICL's primary failure condition is structurally predictable from its design. On tasks where factual consistency is tightly coupled to surface-level lexical overlap between evidence and query most notably HaluEval QA, where each instance includes an explicit supporting knowledge snippet the geometry-aware geometric correction adds limited marginal value over lexical retrieval. In this structured evidence-matching regime, BM25's term-frequency matching already captures the task-relevant signal directly, and the lower-dimensional prototype space learned by GA-ICL does not meaningfully reorganize the geometry beyond what lexical overlap already provides. This boundary condition is constructive rather than limiting: it characterizes precisely the settings where GA-ICL is most valuable, namely open-ended dialogue and summarization tasks where factual consistency requires reasoning over longer context without explicit evidence anchoring. These are also the settings where deployed LLMs are most prone to hallucination in practice. Notably, this QA limitation diminishes at larger model scales on Phi-14B and Qwen3-32B, GA-ICL outperforms all baselines including on QA, suggesting that richer latent geometries in larger models provide more compressible hallucination-relevant structure even in evidence-grounded settings.

\section{Limitations and Future Works}
GA-ICL introduces a one-time training cost of ~3–4 hours and 640MB GPU memory per dataset-model pair, which, while substantially lower than fine-tuning, is an overhead absent from purely heuristic methods. Additionally, the current framework does not support adaptive prototype updates at inference time, which may limit robustness in dynamic or streaming deployment scenarios.
A promising direction for future work is the development of dynamic query-aware prototype selection strategies. In the current framework, prototypes are sampled globally and reused across all test instances, which significantly improves inference efficiency. A natural extension of the dimensionality finding reported here is a systematic study of which attention heads or transformer layers concentrate hallucination-discriminative geometry, which could inform more targeted compression strategies. Another important direction is extending GA-ICL to multimodal hallucination detection settings. While our current work focuses exclusively on textual hallucinations, modern large multimodal models frequently generate hallucinated content in image captioning, visual question answering, and video understanding tasks. Future research could explore whether geometry-aware prototype sampling can effectively model cross-modal latent geometries by jointly leveraging textual and visual representations. This extension may provide a scalable and training-light framework for improving factual reliability in multimodal generative systems.
% While our results demonstrate majority improvements across multiple tasks and model families, the current evaluation is limited to language models with up to 8B parameters, and it remains an open question how manifold based prototype sampling scales to substantially larger models (e.g., 70B+). Additionally, our approach introduces a one time training cost for a lightweight projection head and requires local manifold construction via PCA, which adds computational overhead compared to purely heuristic retrieval methods, though this cost remains significantly lower than fine tuning. Finally, while demonstration selection is deterministic and robust under temperature perturbations, our evaluation does not explore example level uncertainty estimates or adaptive prototype updates at inference time, which could further improve robustness in dynamic or streaming scenarios.

% \section*{Acknowledgments}
% The authors were partially supported by the US National Science Foundation under
% awards IIS-2520978, GEO/RISE-5239902, the Office of Naval Research Award N00014-
% 23-1-2007, DOE (ASCR) Award DE-SC0026052, and the DARPA D24AP00325-00. Approved for public release; distribution is unlimited.

\bibliography{custom}

\appendix

\section{Prompt Templates}
\label{sec:prompt_template}
The following is the prompt template that we used in our experiments to combine the sampled ICL demonstrations with the test query, all the models and baseline methods, including ours, follow the same prompt template.

\begin{lstlisting}[caption={ICL prompt construction for Halueval QA task}, label={lst:icl_prompt}]
def build_icl_prompt(knowledges, questions, answers, few_shots=None):
    prompt = (
        "You are an unbiased document-grounded fact checker. "
        "You are provided a Knowledge, a question based on the knowledge and a answer based on the question. "
        "Based on the provided knowledge for the question identify if the corresponding answer is hallucinated or not.\n"
        "Hallucination response: 'yes' = hallucinated, 'no' = not hallucinated.\n\n"
    )

    if few_shots:
        for know, ques, ans, label in few_shots:
            prompt += (
                f"Knowledge: {know}\n"
                f"Question: {ques}\n"
                f"Answer: {ans}\n"
                f"Hallucination response: [BEGIN]{label}[DONE]\n\n"
            )

    prompt += (
        f"Knowledge: {knowledges}\n"
        f"question: {questions}\n"
        f"Answer: {answers}\n"
        "Hallucination response: [BEGIN]"
    )

    return prompt
\end{lstlisting}

\begin{lstlisting}[caption={ICL prompt construction for HaluEval Dialogue task}, label={lst:icl_dialogue_prompt}]
def build_icl_prompt(knowledges, diag_hists, responses, few_shots=None):
    prompt = (
        "You are an unbiased document-grounded fact checker. "
        "You are provided a Knowledge, a dialogue history based on the knowledge and a response "
        "based on the Knowledge and the dialogue history. "
        "Based on the provided knowledge and dialogue history identify if the corresponding "
        "response is hallucinated or not.\n"
        "Hallucination response: 'yes' = hallucinated, 'no' = not hallucinated.\n\n"
    )

    if few_shots:
        for know, diag_hist, res, label in few_shots:
            prompt += (
                f"Knowledge: {know}\n"
                f"Dialogue history: {diag_hist}\n"
                f"Response: {res}\n"
                f"Hallucination response: [BEGIN]{label}[DONE]\n\n"
            )

    prompt += (
        f"Knowledge: {knowledges}\n"
        f"Dialogue history: {diag_hists}\n"
        f"Response: {responses}\n"
        "Hallucination response: [BEGIN]"
    )

    return prompt
\end{lstlisting}

\begin{lstlisting}[caption={ICL prompt construction for HaluEval Summarization task}, label={lst:icl_summary_prompt}]
def build_icl_prompt(documents, summaries, few_shots=None):
    prompt = (
        "You are an unbiased document-grounded fact checker. "
        "Identify if the corresponding summary for the given document "
        "is hallucinated or not.\n"
        "Hallucination response: 'yes' = hallucinated, "
        "'no' = not hallucinated.\n\n"
    )

    if few_shots:
        for doc, summary, label in few_shots:
            prompt += (
                f"Document: {doc}\n"
                f"Summary: {summary}\n"
                f"Hallucination response: [BEGIN]{label}[DONE]\n\n"
            )

    prompt += (
        f"Document: {documents}\n"
        f"Summary: {summaries}\n"
        "Hallucination response: [BEGIN]"
    )

    return prompt
\end{lstlisting}

\begin{lstlisting}[caption={ICL prompt construction for FEVER dataset}, label={lst:icl_claim_prompt}]
def build_icl_prompt(test_problem, few_shots=None):
    prompt = ""
    if few_shots:
        for prob, sol in few_shots:
            prompt += (
                "You are an unbiased fact checker. "
                "Classify the following claim as supported if it is valid "
                "or refuted if it is invalid:\n"
                f"Claim: {prob}\n"
                "[BEGIN]\n"
                f"{sol}\n"
                "[DONE]\n\n"
            )

    prompt += (
        "You are an unbiased fact checker. "
        "Classify the following claim as supported if it is valid "
        "or refuted if it is invalid:\n"
        f"Claim: {test_problem}\n"
        "[BEGIN]\n"
    )

    return prompt
\end{lstlisting}

\section{Related Works}
\subsection{Hallucination detection}
Identifying hallucinations in large language models is essential for maintaining the trustworthiness and reliability of their generated outputs, particularly in settings where factual accuracy is critical. SelfCheckGPT~\cite{manakul2023selfcheckgptzeroresourceblackboxhallucination} introduces a zero-resource, black-box framework for hallucination detection that operates without access to external knowledge bases. The method is grounded in the observation that when a model has sufficient familiarity with a topic, it tends to produce internally consistent factual statements across multiple generations, whereas responses generated for less familiar content are more likely to exhibit inconsistencies or unsupported claims. The FACTOR ~\cite{muhlgay2024generatingbenchmarksfactualityevaluation} framework evaluates factuality by transforming a factual corpus into controlled multiple-choice benchmarks, generating plausible false alternatives and measuring whether a model assigns higher likelihood to true facts, which enables scalable assessment of factual accuracy without requiring external validation at generation time. 

FActScore ~\cite{min2023factscorefinegrainedatomicevaluation} introduces a fine-grained evaluation metric by decomposing long-form text into atomic facts and computing the percentage of these facts supported by a trusted knowledge source, with an automated estimator that approximates human judgments efficiently. FACTOID ~\cite{rawte2024factoidfactualentailmenthallucination} reframes hallucination detection as a factual entailment problem, jointly identifying whether text is factually supported and localizing specific spans that contain errors, thereby improving upon traditional entailment models for fine-grained factual verification. Finally, FacTool ~\cite{chern2023factoolfactualitydetectiongenerative} builds a task-agnostic, tool-augmented framework that extracts claims from generated outputs and verifies them using appropriate external tools such as search engines or execution environments, grounding factuality judgments in collected evidence rather than model-internal signals alone.

In~\cite{kalai2025languagemodelshallucinate}, the authors argue that a primary factor contributing to hallucinated yet seemingly plausible outputs in large language models is the mismatch between training and inference objectives. Specifically, these objectives tend to incentivize confident guessing rather than explicit acknowledgment of uncertainty, which in turn complicates the identification of the underlying causes of hallucinations. To address the generation of hallucinated content, a wide range of mitigation strategies have been proposed~\cite{sahoo-etal-2024-comprehensive}. These approaches include, but are not limited to, data augmentation and manipulation techniques ~\cite{peng2023checkfactstryagain}, ~\cite{chuang2024doladecodingcontrastinglayers}, alignment and fine-tuning strategies ~\cite{chen2023purrefficientlyeditinglanguage}, ~\cite{zhang2024knowledgealignmentproblembridging}, as well as prompt engineering based methods ~\cite{varshney2023stitchtimesavesnine}, ~\cite{luo2023zeroresourcehallucinationpreventionlarge}, ~\cite{elaraby2023haloestimationreductionhallucinations}.

\section{Performance Considerations}
\label{sec:fine_tuning}
A popular approach to mitigating hallucinations in LLMs is fine-tuning, which modifies the underlying model weights to align with factual distributions. While effective, fine-tuning requires significant GPU memory and multiple training iterations. In contrast, our proposed manifold based sampling framework operates as a lightweight retrieval mechanism that achieves comparable performance gains using a fraction of the resources. By training only a thin projection head $h_\theta$ (see Table~\ref{tab:transformation_architecture}) and keeping the base model frozen, our method reduces the training memory footprint by several orders of magnitude,  making GA-ICL suitable for domain-specific hallucination where fine-tuning is impractical. 

We evaluate the performance of GA-ICL against a fine-tuned baseline. All experiments are conducted on Vicuna-7B using A6000 RTX GPUs with 48 GB of memory. Fine-tuning the 16-bit model requires two GPUs, while all other fine-tuning configurations are run on a single GPU. For each model, we attached LoRA adapters to the last 3 to 5 layers, including the highest performing configuration in Table~\ref{tab:efficiency_comparison}. A summary of experiment configurations and computational results is provided in Table~\ref{tab:efficiency_comparison}.

\begin{table}[t]
\centering
\small
\begin{tabular}{lccc}
\toprule
Method & Memory (GB) & Time (hrs) & Acc \\
\midrule
LoRA-SFT, 4-bit  & 24 & 2.6 & 0.71 \\
LoRA-SFT, 8-bit & 40 & 3.6 & 0.74 \\
LoRA-SFT, 16-bit & 72 & 4.1 & 0.73 \\
GA-ICL (ours)   & 0.64     & 3.4 & 0.71 \\
\bottomrule
\end{tabular}
\caption{Comparison of computational cost and performance between supervised fine-tuning (SFT) and GA-ICL for Vicuna-7B model. GA-ICL achieves comparable performance with orders-of-magnitude lower GPU memory usage.}
\label{tab:efficiency_comparison}
\end{table}

\begin{figure*}[t]
    \centering
    
    % Qwen3-4B Plot
    \begin{subfigure}{0.48\textwidth}
        \centering
        \begin{tikzpicture}
            \begin{axis}[
                width=\textwidth,
                height=6cm,
                xlabel={No. of Demonstrations},
                ylabel={Accuracy},
                title={Accuracy vs no. of Demonstrations for Qwen3-4B},
                legend pos=south east,
                legend style={font=\small, fill opacity=0.7, text opacity=1},
                grid=major,
                mark size=2.5pt,
                xtick={2,4,6,8,10,12},
                xmin=1.5, xmax=12.5,
                ymin=0.65, ymax=0.85,
            ]
            
            % Halueval Dialogue task
            \addplot[mark=*, blue, thick] coordinates {
                (2, 0.72) (4, 0.737) (6, 0.7401) (8, 0.7492) (10, 0.7512) (12, 0.7467)
            };
            \addlegendentry{Dialogue Task}
            
            % Halueval QA task
            \addplot[mark=x, red, thick] coordinates {
                (2, 0.705) (4, 0.787) (6, 0.7922) (8, 0.8039) (10, 0.8154) (12, 0.8058)
            };
            \addlegendentry{QA Task}
            
            \end{axis}
        \end{tikzpicture}
        \caption{Qwen3-4B}
        \label{fig:qwen3_accuracy}
    \end{subfigure}
    \hfill
    % Falcon3-3B Plot
    \begin{subfigure}{0.48\textwidth}
        \centering
        \begin{tikzpicture}
            \begin{axis}[
                width=\textwidth,
                height=6cm,
                xlabel={No. of Demonstrations},
                ylabel={Accuracy},
                title={Accuracy vs no. of Demonstrations for Falcon3-3B},
                legend pos=south east,
                legend style={font=\small, fill opacity=0.7, text opacity=1},
                grid=major,
                mark size=2.5pt,
                xtick={2,4,6,8,10,12},
                xmin=1.5, xmax=12.5,
                ymin=0.48, ymax=0.58,
            ]
            
            % Halueval Dialogue task
            \addplot[mark=*, blue, thick] coordinates {
                (2, 0.5099) (4, 0.51) (6, 0.5432) (8, 0.5648) (10, 0.5695) (12, 0.5578)
            };
            \addlegendentry{Dialogue Task}
            
            % Halueval QA task
            \addplot[mark=x, red, thick] coordinates {
                (2, 0.5051) (4, 0.5021) (6, 0.51) (8, 0.5229) (10, 0.5258) (12, 0.5073)
            };
            \addlegendentry{QA Task}
            
            \end{axis}
        \end{tikzpicture}
        \caption{Falcon3-3B}
        \label{fig:falcon3_accuracy}
    \end{subfigure}
    
    \caption{Performance of GA-ICL under varying numbers of in-context demonstrations}
    \label{fig:model_comparison}
\end{figure*}
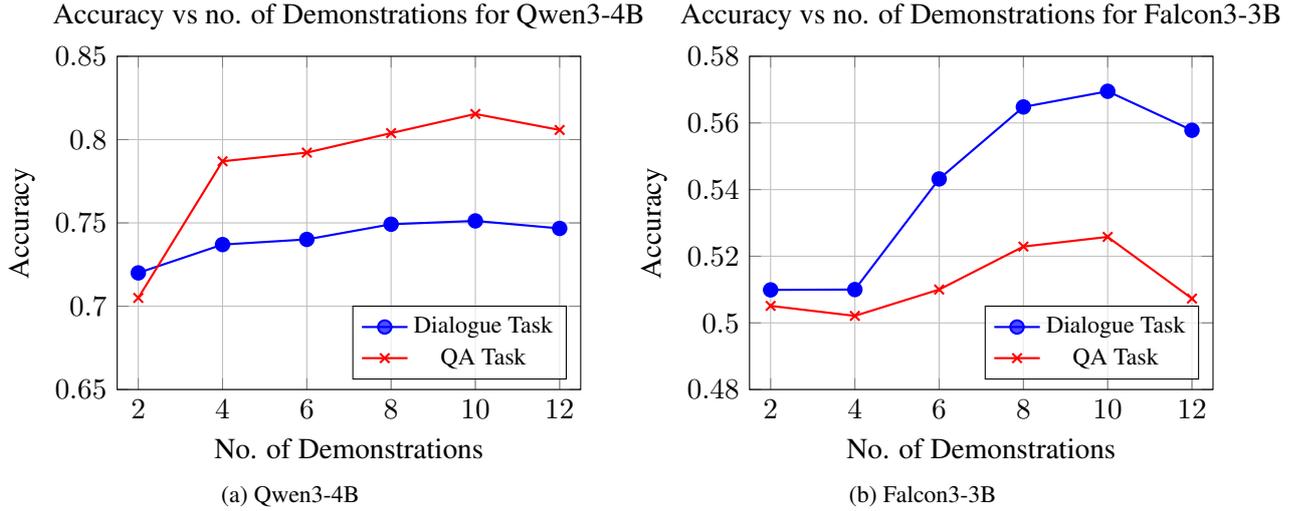

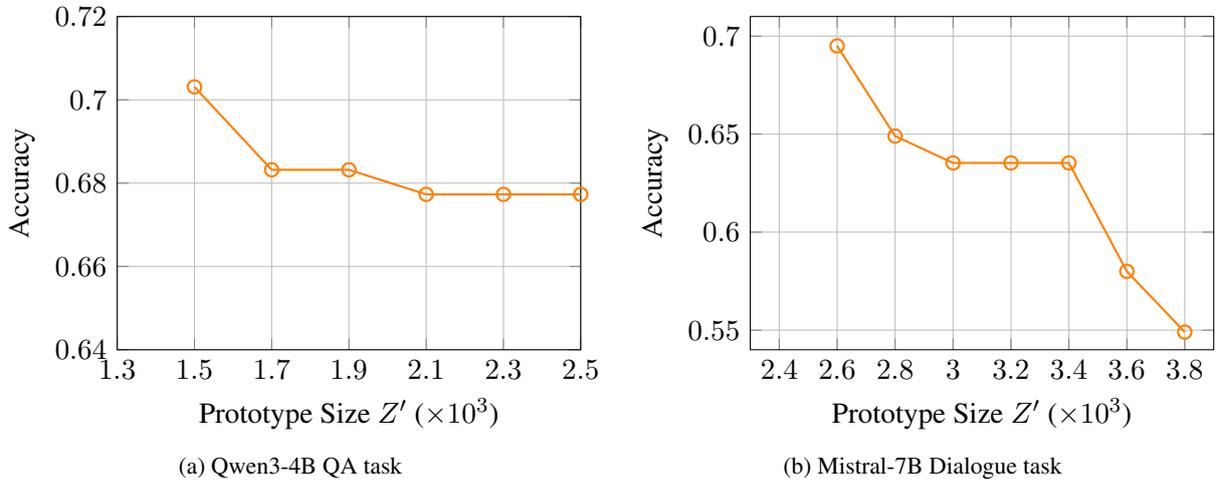
\begin{figure*}[t]
    \centering
    % Qwen3-4B Plot
    \begin{subfigure}[t]{0.32\textwidth}
        \centering
        \begin{tikzpicture}
            \begin{axis}[
                width=0.95\linewidth,
                height=5cm,
                xlabel={Prototype Size $Z'$ ($\times 10^3$)},
                ylabel={Accuracy},
                grid=major,
                mark size=2.5pt,
                xtick={1.5,1.9,2.1,2.5},
                xticklabel style={rotate=45, anchor=east, font=\small},
                xmin=1.3, xmax=2.6,
                ymin=0.64, ymax=0.72,
            ]
            \addplot[mark=o, orange, thick] coordinates {
                (2.5, 0.6773)
                (2.3, 0.6773)
                (2.1, 0.6773)
                (1.9, 0.6832)
                (1.7, 0.6832)
                (1.5, 0.7031)
            };
            \end{axis}
        \end{tikzpicture}
        \caption{Qwen3-4B QA task}
        \label{fig:qwen_latent_size_accuracy}
    \end{subfigure}%
    \hfill
    % Mistral-7B Dialogue Plot
    \begin{subfigure}[t]{0.32\textwidth}
        \centering
        \begin{tikzpicture}
            \begin{axis}[
                width=0.95\linewidth,
                height=5cm,
                xlabel={Prototype Size $Z'$ ($\times 10^3$)},
                ylabel={Accuracy},
                grid=major,
                mark size=2.5pt,
                xtick={2.6,3.0,3.4,3.8},
                xticklabel style={rotate=45, anchor=east, font=\small},
                xmin=2.3, xmax=3.9,
                ymin=0.54, ymax=0.71,
            ]
            \addplot[mark=o, orange, thick] coordinates {
                (3.8, 0.549)
                (3.6, 0.58)
                (3.4, 0.6353)
                (3.2, 0.6353)
                (3.0, 0.6353)
                (2.8, 0.649)
                (2.6, 0.695)
            };
            \end{axis}
        \end{tikzpicture}
        \caption{Mistral-7B Dialogue task}
        \label{fig:mistral_dialogue_latent_size_accuracy}
    \end{subfigure}%
    \hfill
    % Mistral-7B QA Plot
    \begin{subfigure}[t]{0.32\textwidth}
        \centering
        \begin{tikzpicture}
            \begin{axis}[
                width=0.95\linewidth,
                height=5cm,
                xlabel={Prototype Size $Z'$ ($\times 10^3$)},
                ylabel={Accuracy},
                grid=major,
                mark size=2.5pt,
                xtick={2.6,3.0,3.4,3.8},
                xticklabel style={rotate=45, anchor=east, font=\small},
                xmin=2.3, xmax=3.9,
                ymin=0.52, ymax=0.66,
            ]
            \addplot[mark=o, orange, thick] coordinates {
                (3.8, 0.53)
                (3.6, 0.58)
                (3.4, 0.567)
                (3.2, 0.591)
                (3.0, 0.598)
                (2.8, 0.612)
                (2.6, 0.634)
            };
            \end{axis}
        \end{tikzpicture}
        \caption{Mistral-7B QA task}
        \label{fig:mistral_qa_latent_size_accuracy}
    \end{subfigure}
    \caption{Accuracy vs Prototype Size $Z'$}
    \label{fig:prototype_size_comparison}
\end{figure*}

\subsection{Comparison methods}
We compare our approach against five baseline sampling methods. \textbf{KNN} selects the most similar example(s) from the dataset for each query based on cosine similarity. \textbf{Clustering} chooses examples that are closest to the class centroid within each group. We note that clustering-based selection often performs near chance level in hallucination detection settings, reflecting the limited alignment between global semantic clusters and factual consistency labels. The \textbf{Perplexity} ~\cite{gonen-etal-2023-demystifying} based method follows prior work on perplexity driven selection, where all input fields except the label are concatenated into a text prompt and passed through the corresponding LLM to compute perplexity scores. Prompts with the lowest perplexity values are then selected as in context learning (ICL) demonstrations. \textbf{BM25} is a lexical similarity based retrieval method that ranks candidate examples using term frequency inverse document frequency (TF–IDF) statistics and selects the top scoring examples as demonstrations. \textbf{SA-ICL} (Self-Adaptive ICL) ~\cite{wu-etal-2023-self} is based on prior self-adaptive ICL work, where the top-$k$ relevant demonstrations for each test query are first retrieved using similarity measures and subsequently reordered to obtain an arrangement that minimizes the code length of the label $Y$, following the Minimum Description Length (MDL) principle.
\textbf{AdaICL} ~\cite{mavromatis2023examplesannotateincontextlearning} identifies "hard" examples ones the LLM is most uncertain about (low max confidence over labels) then uses a greedy Maximum Coverage algorithm over their SBERT-based neighborhood graph to select a diverse, representative annotation budget. At inference, a k-NN retriever pulls the most similar annotated examples as ICL demonstrations for each test query. \textbf{TopK+ConE} ~\cite{peng2024revisitingdemonstrationselectionstrategies} first retrieves the M most semantically similar candidates to a test query via SBERT cosine similarity (TopK), then reranks them by computing the conditional entropy of the test input under the LLM itself selecting whichever demonstrations minimise the model's uncertainty about the test input, making the choice both data- and model-dependent.

\section{Extended Evaluations}
\label{sec:entended_eval}
We compare against learning-based selection methods only at this scale, as heuristic baselines such as BM25 and KNN do not require per-model training and their relative performance characteristics on smaller models are already established in Table ~\ref{tab:main_results}; the primary question at larger scale is whether GA-ICL's learned geometry remains competitive opnst the strongest adaptive baselines. The results presented in Table ~\ref{tab:large_model_results} demonstrate that GA-ICL scales effectively to larger models and consistently outperforms the recent baseline methods.

Following the prototype-dimensionality ablation discussed in Section ~\ref{sec:details_and_results}, we observed that aggressive manifold compression substantially improved performance on Mistral-7B QA (Figure ~\ref{fig:mistral_qa_latent_size_accuracy}. The updated results therefore use the optimized prototype dimensionality identified through the manifold-reduction study.

\section{Inference comparison of time taken and memory consumed}
\label{sec: inference_table}

\begin{table}[h]
\centering
\renewcommand{\arraystretch}{1.0}
\setlength{\tabcolsep}{3pt}
\footnotesize
\begin{tabular}{@{}llcc@{}}
\toprule
\textbf{Task} & \textbf{Method} & \textbf{Time} & \textbf{Mem (MB)} \\
\midrule
\multirow{4}{*}{FEVER}
 & BM25       & 50 min        & 6584 \\
 & KNN        & 1 hr 30 min   & 6584 \\
 & Perplexity & 55 min        & 9392 \\
 & GA-ICL     & \textbf{25 min} & \textbf{6574} \\
\midrule
\multirow{5}{*}{\shortstack[l]{HaluEval\\Summ.}}
 & BM25       & 7 hr          & 8196  \\
 & KNN        & 3 hr          & 9260  \\
 & Perplexity & 2.3 hr        & 31178 \\
 & SA-ICL     & 3 hr          & 13574 \\
 & GA-ICL     & \textbf{1 hr} & \textbf{7900} \\
\midrule
\multirow{5}{*}{\shortstack[l]{HaluEval\\QA}}
 & BM25       & 1 hr          & 6769  \\
 & KNN        & 1 hr 20 min   & 6714  \\
 & Perplexity & 55 min        & 29116 \\
 & SA-ICL     & 1 hr          & 9528  \\
 & GA-ICL     & \textbf{25 min} & \textbf{6685} \\
\midrule
\multirow{5}{*}{\shortstack[l]{HaluEval\\Diag.}}
 & BM25       & 1 hr          & 6778  \\
 & KNN        & 1 hr 20 min   & 6752  \\
 & Perplexity & 55 min        & 30051 \\
 & SA-ICL     & 1 hr          & 9643  \\
 & GA-ICL     & \textbf{25 min} & \textbf{6694} \\
\bottomrule
\end{tabular}
\caption{Inference time and memory usage across tasks and methods (Falcon-3-3B).}
\label{tab:falcon-benchmarks}
\end{table}

\section{Implementation details}
\label{sec:imple_details}

The network $h_\theta$, which is crucial in our method to sample the ICL demonstrations, is trained for 200 epochs on the training dataset $D_{F_t}$. Training utilizes two independent Adam optimizers: one for the network parameters and another for the proxy parameters. Both optimizers are initialized with a learning rate of \texttt{1e-3}, combined with a scheduler that decays the learning rate by a factor of $\eta_t = 0.97$. The dimensionality of the encoded vector $Z$ is determined by the underlying Large Language Model ($M$). A mini-batch size of 128 samples is maintained throughout training.  

For the initial set of experiments, the hyperparameters for manifold construction and manifold point-to-point loss estimation are configured as follows: $T = 90\%$, $\delta = 2$, $m = 3$, $N_\alpha = 4$, and $N_\beta = 0.5$. The momentum constant for updating $\theta_m$ is set to $\mu = 0.99$. For Proxy Anchor loss, we employ $\alpha = 32$ and $\epsilon = 0.1$. These settings serve as the baseline configuration; subsequently, an ablation study is conducted on the above parameters for LLMs that exhibited comparatively lower performance than competing methods.

All experiments were conducted on an NVIDIA RTX A6000 GPU. Training the lightweight neural network $h_\theta$ requires approximately 640 MB of GPU memory and about 3-4 hours of training time.

\begin{figure*}[t]
    \centering
    
    % QA Accuracy Plots
    \begin{subfigure}{0.48\textwidth}
        \centering
        \begin{tikzpicture}
            \begin{axis}[
                width=\textwidth,
                height=5cm,
                xlabel={Temperature},
                ylabel={Accuracy},
                title={Qwen3 HaluEval QA Accuracy vs Temperature},
                legend pos=south west,
                legend style={font=\tiny, fill opacity=0.7, text opacity=1},
                grid=major,
                mark size=2pt,
                cycle list name=color list,
            ]
            % Qwen ProTS
            \addplot[mark=o, blue, thick] coordinates {
                (0.1, 0.7021) (0.2, 0.7023) (0.3, 0.7021) (0.4, 0.7025) (0.5, 0.7012)
                (0.6, 0.7017) (0.7, 0.7031) (0.8, 0.7062) (0.9, 0.7041) (1.0, 0.7057)
            };
            \addlegendentry{GA-ICL}
            
            % Qwen Self-Consistency
            \addplot[mark=o, red, thick] coordinates {
                (0.1, 0.71) (0.2, 0.73) (0.3, 0.69) (0.4, 0.73) (0.5, 0.67)
                (0.6, 0.71) (0.7, 0.72) (0.8, 0.75) (0.9, 0.71) (1.0, 0.73)
            };
            \addlegendentry{SA-ICL}
            
            % Qwen BM25
            \addplot[mark=o, green!60!black, thick] coordinates {
                (0.1, 0.69) (0.2, 0.71) (0.3, 0.697) (0.4, 0.71) (0.5, 0.69)
                (0.6, 0.65) (0.7, 0.71) (0.8, 0.69) (0.9, 0.71) (1.0, 0.65)
            };
            \addlegendentry{BM25}
            \end{axis}
        \end{tikzpicture}
        \caption{Qwen3 HaluEval QA}
        \label{fig:qa_qwen}
    \end{subfigure}
    \hfill
    \begin{subfigure}{0.48\textwidth}
        \centering
        \begin{tikzpicture}
            \begin{axis}[
                width=\textwidth,
                height=5cm,
                xlabel={Temperature},
                ylabel={Accuracy},
                title={Falcon3 HaluEval QA Accuracy vs Temperature},
                legend pos=south west,
                legend style={font=\tiny, fill opacity=0.7, text opacity=1},
                grid=major,
                mark size=2pt,
            ]
            % Falcon ProTS
            \addplot[mark=o, blue, thick] coordinates {
                (0.1, 0.499) (0.2, 0.501) (0.3, 0.5067) (0.4, 0.5034) (0.5, 0.508)
                (0.6, 0.510) (0.7, 0.508) (0.8, 0.51) (0.9, 0.506) (1.0, 0.503)
            };
            \addlegendentry{GA-ICL}
            
            % Falcon Self-Consistency
            \addplot[mark=o, red, thick] coordinates {
                (0.1, 0.50) (0.2, 0.51) (0.3, 0.48) (0.4, 0.53) (0.5, 0.52)
                (0.6, 0.53) (0.7, 0.52) (0.8, 0.49) (0.9, 0.53) (1.0, 0.50)
            };
            \addlegendentry{SA-ICL}
            
            % Falcon BM25
            \addplot[mark=o, green!60!black, thick] coordinates {
                (0.1, 0.49) (0.2, 0.52) (0.3, 0.502) (0.4, 0.51) (0.5, 0.49)
                (0.6, 0.51) (0.7, 0.49) (0.8, 0.52) (0.9, 0.52) (1.0, 0.49)
            };
            \addlegendentry{BM25}
            \end{axis}
        \end{tikzpicture}
        \caption{Falcon3 HaluEval QA}
        \label{fig:qa_falcon}
    \end{subfigure}
    
    \caption{Accuracy vs Temperature comparison for Qwen3 and Falcon3 models on HaluEval QA dataset}
    \label{fig:qa_accuracy}
\end{figure*}
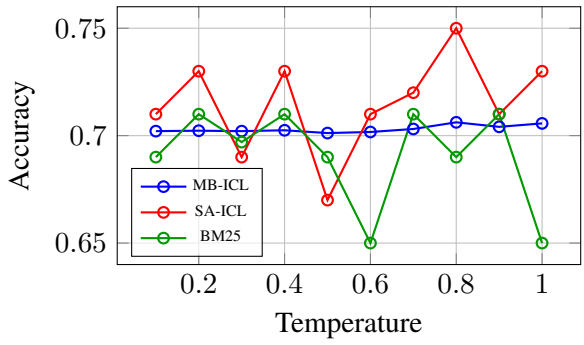
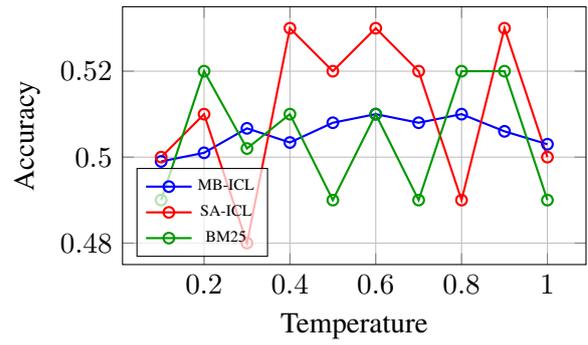

\begin{figure*}[t]
    \centering
    
    % Diagnostic Accuracy Plots
    \begin{subfigure}{0.48\textwidth}
        \centering
        \begin{tikzpicture}
            \begin{axis}[
                width=\textwidth,
                height=5cm,
                xlabel={Temperature},
                ylabel={Accuracy},
                title={Qwen3 HaluEval Dialogue Accuracy vs Temperature},
                legend pos=south west,
                legend style={font=\tiny, fill opacity=0.7, text opacity=1},
                grid=major,
                mark size=2pt,
            ]
            % Qwen ProTS
            \addplot[mark=o, blue, thick] coordinates {
                (0.1, 0.7155) (0.2, 0.715) (0.3, 0.7134) (0.4, 0.7169) (0.5, 0.7138)
                (0.6, 0.7125) (0.7, 0.7113) (0.8, 0.712) (0.9, 0.709) (1.0, 0.7089)
            };
            \addlegendentry{GA-ICL}
            
            % Qwen Self-Consistency
            \addplot[mark=o, red, thick] coordinates {
                (0.1, 0.72) (0.2, 0.71) (0.3, 0.69) (0.4, 0.719) (0.5, 0.712)
                (0.6, 0.68) (0.7, 0.712) (0.8, 0.701) (0.9, 0.69) (1.0, 0.7)
            };
            \addlegendentry{SA-ICL}
            
            % Qwen BM25
            \addplot[mark=o, green!60!black, thick] coordinates {
                (0.1, 0.73) (0.2, 0.72) (0.3, 0.68) (0.4, 0.705) (0.5, 0.70)
                (0.6, 0.698) (0.7, 0.70) (0.8, 0.72) (0.9, 0.71) (1.0, 0.68)
            };
            \addlegendentry{BM25}
            \end{axis}
        \end{tikzpicture}
        \caption{Qwen3 HaluEval Dialogue task}
        \label{fig:diag_qwen}
    \end{subfigure}
    \hfill
    \begin{subfigure}{0.48\textwidth}
        \centering
        \begin{tikzpicture}
            \begin{axis}[
                width=\textwidth,
                height=5cm,
                xlabel={Temperature},
                ylabel={Accuracy},
                title={Falcon3 HaluEval Dialogue Accuracy vs Temperature},
                legend pos=south west,
                legend style={font=\tiny, fill opacity=0.7, text opacity=1},
                grid=major,
                mark size=2pt,
            ]
            % Falcon ProTS
            \addplot[mark=o, blue, thick] coordinates {
                (0.1, 0.5014) (0.2, 0.5038) (0.3, 0.506) (0.4, 0.5107) (0.5, 0.5064)
                (0.6, 0.503) (0.7, 0.4989) (0.8, 0.5002) (0.9, 0.5082) (1.0, 0.4983)
            };
            \addlegendentry{GA-ICL}
            
            % Falcon Self-Consistency
            \addplot[mark=o, red, thick] coordinates {
                (0.1, 0.51) (0.2, 0.50) (0.3, 0.49) (0.4, 0.51) (0.5, 0.512)
                (0.6, 0.51) (0.7, 0.47) (0.8, 0.51) (0.9, 0.52) (1.0, 0.49)
            };
            \addlegendentry{SA-ICL}
            
            % Falcon BM25
            \addplot[mark=o, green!60!black, thick] coordinates {
                (0.1, 0.50) (0.2, 0.513) (0.3, 0.5) (0.4, 0.48) (0.5, 0.503)
                (0.6, 0.50) (0.7, 0.51) (0.8, 0.50) (0.9, 0.50) (1.0, 0.51)
            };
            \addlegendentry{BM25}
            \end{axis}
        \end{tikzpicture}
        \caption{Falcon3 HaluEval Dialogue task}
        \label{fig:diag_falcon}
    \end{subfigure}
    
    \caption{ Accuracy vs Temperature comparisons for Qwen3 and Falcon3 models on HaluEval Dialogue dataset}
    \label{fig:diag_accuracy}
\end{figure*}
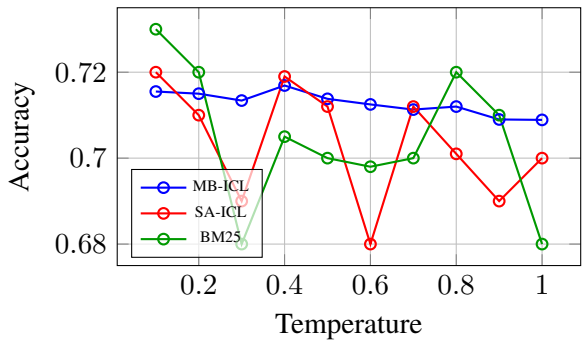
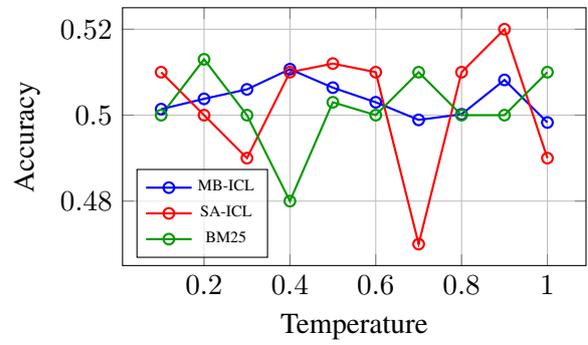

\begin{figure*}[t]
    \centering
    \begin{subfigure}[b]{0.48\textwidth}
        \centering
        \begin{tikzpicture}[scale=0.8]
            \begin{axis}[
                xlabel={$\mu$},
                ylabel={Accuracy},
                grid=major,
                width=\textwidth,
                height=0.7\textwidth,
                mark size=2pt,
                legend pos=north east,
                ymin=0.52,
                ymax=0.6
            ]
            \addplot[blue, mark=*, thick] coordinates {
                (0.3, 0.5697)
                (0.5, 0.5678)
                (0.9, 0.5681)
                (0.99, 0.5697)
                (0.999, 0.5591)
                (1, 0.5621)
            };
            \end{axis}
        \end{tikzpicture}
        \caption{$\mu$ vs Accuracy}
        \label{fig:subfig1}
    \end{subfigure}
    \hfill
    \begin{subfigure}[b]{0.48\textwidth}
        \centering
        \begin{tikzpicture}[scale=0.8]
            \begin{axis}[
                xlabel={$m$},
                ylabel={Accuracy},
                grid=major,
                width=\textwidth,
                height=0.7\textwidth,
                mark size=2pt,
                legend pos=north east,
                ymin=0.52,
                ymax=0.6
            ]
            \addplot[red, mark=square*, thick] coordinates {
                (1, 0.5697)
                (2, 0.5666)
                (3, 0.5697)
                (4, 0.5631)
                (5, 0.5697)
                (6, 0.5623)
                (7, 0.5697)
                (8, 0.5597)
            };
            \end{axis}
        \end{tikzpicture}
        \caption{$m$ vs Accuracy}
        \label{fig:subfig2}
    \end{subfigure}
    
    \vspace{0.5cm}
    
    \begin{subfigure}[b]{0.48\textwidth}
        \centering
        \begin{tikzpicture}[scale=0.8]
            \begin{axis}[
                xlabel={$N_\beta$},
                ylabel={Accuracy},
                grid=major,
                width=\textwidth,
                height=0.7\textwidth,
                mark size=2pt,
                legend pos=north east,
                ymin=0.52,
                ymax=0.6
            ]
            \addplot[green!60!black, mark=triangle*, thick] coordinates {
                (0.5, 0.5697)
                (1, 0.5666)
                (1.5, 0.5722)
                (2, 0.5708)
                (2.5, 0.5697)
                (3, 0.568)
            };
            \end{axis}
        \end{tikzpicture}
        \caption{$N_\beta$ vs Accuracy}
        \label{fig:subfig3}
    \end{subfigure}
    \hfill
    \begin{subfigure}[b]{0.48\textwidth}
        \centering
        \begin{tikzpicture}[scale=0.8]
            \begin{axis}[
                xlabel={$N_\alpha$},
                ylabel={Accuracy},
                grid=major,
                width=\textwidth,
                height=0.7\textwidth,
                mark size=2pt,
                legend pos=north east,
                ymin=0.52,
                ymax=0.6
            ]
            \addplot[orange, mark=diamond*, thick] coordinates {
                (1, 0.5697)
                (2, 0.568)
                (3, 0.5697)
                (4, 0.5722)
                (5, 0.5666)
                (6, 0.5681)
                (7, 0.5697)
                (8, 0.5697)
            };
            \end{axis}
        \end{tikzpicture}
        \caption{$N_\alpha$ vs Accuracy}
        \label{fig:subfig4}
    \end{subfigure}
    
    \caption{Hyperparameter tuning for Qwen3-4B on Summarization task}
    \label{fig:hyper_param_overall}
\end{figure*}
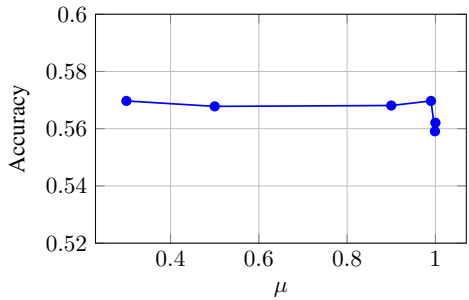
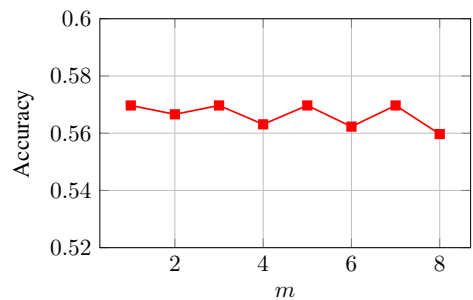
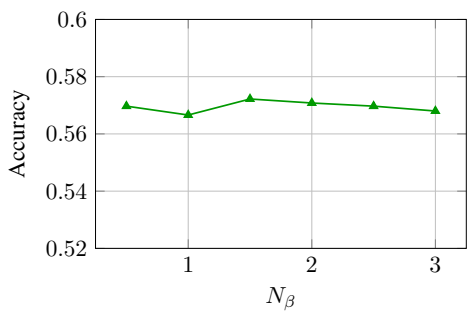
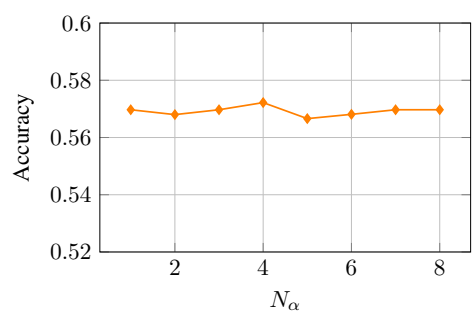

\section{AI usage}
We have used AI tools(Chatgpt) for writing assistance. We did not use it during any of the ideation or experimentation phase.

\end{document}